\documentclass[10pt,twocolumn,letterpaper]{article}
\usepackage[dvipsnames]{xcolor}

\usepackage{cvpr}              

\usepackage{times}
\usepackage{epsfig}
\usepackage{graphicx}
\usepackage{amsmath}
\usepackage{amssymb}
\usepackage{booktabs}
\usepackage{cite}
\usepackage{color}
\usepackage{times}
\usepackage{epsfig}
\usepackage{graphicx}
\usepackage{graphics}
\usepackage{amsmath}
\usepackage{amssymb}
\usepackage{multirow}
\usepackage{mathrsfs}
\usepackage{rotating}
\usepackage{bbm}
\usepackage{verbatim}
\usepackage{array}
\usepackage{tabularx}
\usepackage{url}
\usepackage{marvosym}
\usepackage{xcolor}
\definecolor{myblue}{RGB}{0,0,128}
\definecolor{mygray}{RGB}{128, 128, 128}
\definecolor{cvprblue}{rgb}{0.21,0.49,0.74}
\usepackage{colortbl}
\PassOptionsToPackage{table}{xcolor}
\usepackage[table]{xcolor}
\usepackage[pagebackref,breaklinks,colorlinks,citecolor=cvprblue]{hyperref}
\usepackage[capitalize]{cleveref}
\crefname{section}{Sec.}{Secs.}
\Crefname{section}{Section}{Sections}
\Crefname{table}{Table}{Tables}
\crefname{table}{Tab.}{Tabs.}

\def\paperID{3851} 
\def\confName{CVPR}
\def\confYear{2024}

\begin{document}

\title{
Frequency-Adaptive Dilated Convolution for Semantic Segmentation
}

\author{
Linwei Chen$^1$ \qquad \qquad Lin Gu$^{2,3}$ \qquad \qquad Dezhi Zheng$^{1}$ \qquad \qquad Ying Fu$^{1*}$ \\
$^1$Beijing Institute of Technology \qquad $^2$RIKEN \qquad $^3$The University of Tokyo \\
{\tt\small chenlinwei@bit.edu.cn \qquad lin.gu@riken.jp \qquad zhengdezhi@bit.edu.cn\qquad fuying@bit.edu.cn}
}
\maketitle
\let\thefootnote\relax\footnotetext{*Corresponding Author}
\begin{abstract}
Dilated convolution, which expands the receptive field by inserting gaps between its consecutive elements, is widely employed in computer vision. In this study, we propose three strategies to improve individual phases of dilated convolution from the perspective of spectrum analysis. Departing from the conventional practice of fixing a global dilation rate as a hyperparameter, we introduce Frequency-Adaptive Dilated Convolution (FADC), which dynamically adjusts dilation rates spatially based on local frequency components. Subsequently, we design two plug-in modules to directly enhance effective bandwidth and receptive field size.
The Adaptive Kernel (AdaKern) module decomposes convolution weights into low-frequency and high-frequency components, dynamically adjusting the ratio between these components on a per-channel basis. By increasing the high-frequency part of convolution weights, AdaKern captures more high-frequency components, thereby improving effective bandwidth.
The Frequency Selection (FreqSelect) module optimally balances high- and low-frequency components in feature representations through spatially variant reweighting. It suppresses high frequencies in the background to encourage FADC to learn a larger dilation, thereby increasing the receptive field for an expanded scope. Extensive experiments on segmentation and object detection consistently validate the efficacy of our approach.
The code is made publicly available at \href{https://github.com/Linwei-Chen/FADC}{https://github.com/ying-fu/FADC}.

\end{abstract}



\section{Introduction}
\label{sec:intro}
\begin{figure}[t!]
\centering
\scalebox{0.98}{
\begin{tabular}{cc}
\includegraphics[width=0.98\linewidth]{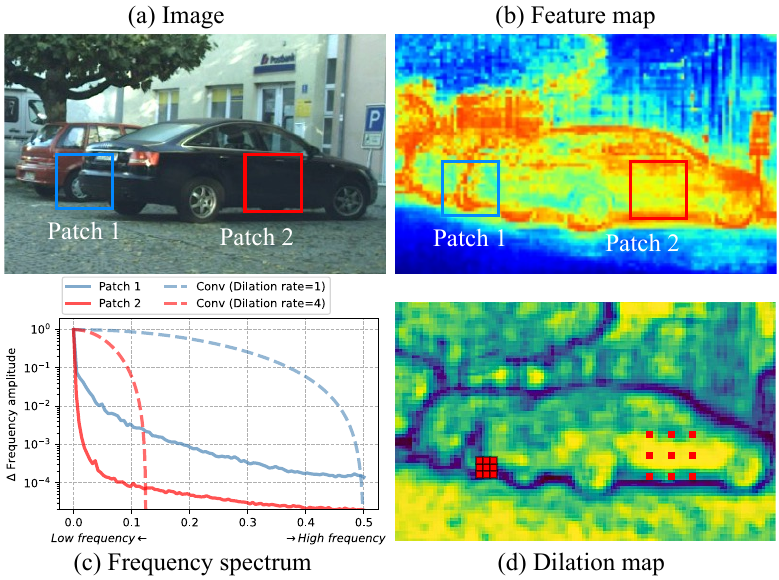} \\
\vspace{-6mm}
\end{tabular}}
\caption{
For the input image (a), its extracted features (b) exhibit spatial variation. Patch 1 contains more high-frequency information, whereas the frequency of patch 2 is predominantly concentrated in low frequency (c). Consequently, assigning a small dilation rate for patch 1 is essential to preserve a high effective bandwidth, while a larger dilation rate, with limited effective bandwidth, is sufficient for patch 2, benefiting the achievement of a larger receptive field (d).
}
\label{fig:intro}
\vspace{-5mm} 
\end{figure}
Dilated convolution inserts gaps between the filter values at a dilation rate (\(D\)) to expand the receptive field without significantly increasing computational load. This technique is widely used in computer vision tasks, such as semantic segmentation~\cite{2017dilated, deeplabv3plus} and object detection~\cite{fasterRCNN2015}.

While effective in expanding the receptive field size with a large dilation rate, it comes at the expense of high-frequency component response~\cite{2017dilated}. Increasing the dilation rate from 1 to \(D\) is equivalent to expanding the convolution kernel through zero-insertion by a factor of \(D\). According to the scaling property of Fourier Transforms~\cite{1987digital, 2013digital}, both the frequency response curve and the bandwidth of the convolution kernel will be scaled to \(\frac{1}{D}\). As illustrated in Figure~\ref{fig:intro}, the bandwidth of the red curve for \(D = 4\) is only a quarter of the one for \(D = 1\) in blue. The reduced bandwidth significantly limits the layer's ability to process high-frequency components. For instance, gridding artifacts occur when a feature map has higher-frequency content than the sampling rate of the dilated convolution~\cite{2017dilated, 2018understanding}.

\begin{figure}[tb!]
\centering
\includegraphics[width=0.98\linewidth]{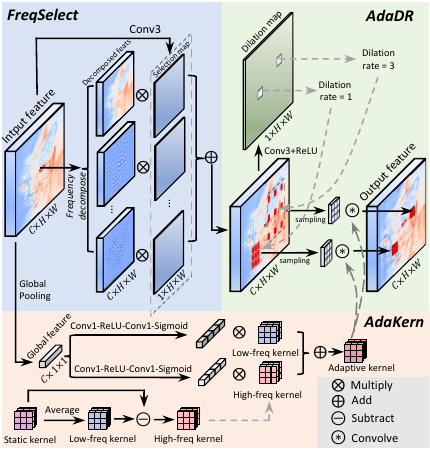}
\vspace{-3mm}   
\caption{
Overview of the proposed Frequency Adaptive Dilated Convolution (FADC). It comprises three strategies: Adaptive Dilation Rate (AdaDR), Adaptive Kernel (AdaKern), and Frequency Selection (FreqSelect).
}
\label{fig:FADC}
\vspace{-5mm} 
\end{figure}


\begin{figure}[tb!]
\centering
\scalebox{0.928}{
\begin{tabular}{cc}
\hspace{-5mm}
\includegraphics[height=0.518\linewidth]{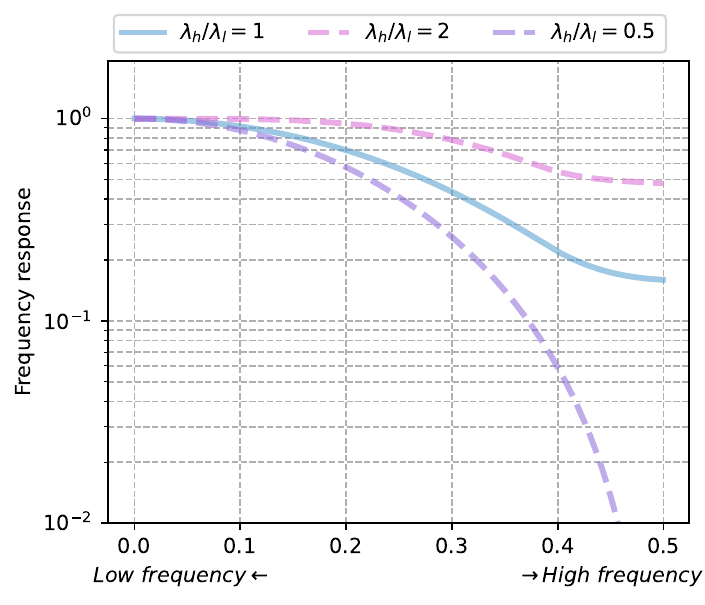}
\hspace{-5mm}
&\includegraphics[height=0.528\linewidth]{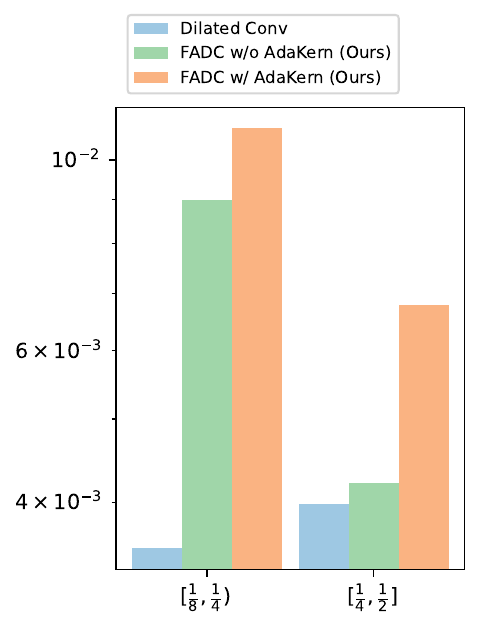} \\
\vspace{-9.18mm}
\end{tabular}}
\caption{
Analysis of AdaKern.
Left: representative frequency response of a convolution kernel, $\lambda_l$, $\lambda_h$ represent dynamic weights for decomposed low/high-frequency kernels. 
Right: the FADC with AdaKern increases the proportion of high-frequency band power within $[\frac{1}{8}, \frac{1}{4})$ and $[\frac{1}{4}, \frac{1}{2}]$ in the extracted feature, indicating an increase in effective bandwidth.
}
\label{fig:adakern}
\vspace{-5mm} 
\end{figure}

In this paper, we introduce Frequency-Adaptive Dilated Convolution (FADC) to enhance dilated convolution through the lens of spectrum analysis. As illustrated in Figure~\ref{fig:FADC}, FADC comprises three key strategies, \ie, Adaptive Dilation Rate (AdaDR), Adaptive Kernel (AdaKern), and Frequency Selection (FreqSelect), aimed at enhancing the individual phases of vanilla dilated convolution. AdaDR spatially adjusts the dilation rate, AdaKern operates on the convolution kernel weights, while FreqSelect directly balances the frequency power of the input feature to encourage the expansion of the receptive field.

Unlike the conventional approach of globally fixing the dilation rate, our AdaDR dynamically assigns dilation rates locally based on the spectrum. For instance, in patch 1 of Figure~\ref{fig:intro}{\color{red}(a)}, where the car boundaries exhibit much high-frequency component (indicated by the blue solid line), AdaDR applies a small dilation rate ($D = 1$) with a broad effective bandwidth (represented by the blue dot curve). 
Conversely, for the car door in patch 2, where the frequency power is predominantly concentrated in the low-frequency domain, AdaDR increases the dilation rate $D$ to 4, as a reduced bandwidth can still encompass a substantial amount of frequency power. The dilation map for these two patches is illustrated in Figure~\ref{fig:intro}{\color{red}(d)}. 
In comparison to the fixed dilation rates (\eg, $D = $1, 2, 4 in~\cite{2017dilated, 2022pcaa}), our AdaDR enhances the theoretical average receptive field size of Figure~\ref{fig:intro} from $\sim$440 to $\sim$1000 pixels.

AdaKern is a plug-in module that manipulates the convolution kernel to optimize the frequency response curve in Figure~\ref{fig:adakern} and enhances the effective bandwidth. As shown in Figure~\ref{fig:adakern}, this module decomposes convolution weights into low-frequency and high-frequency components. This allows us to dynamically manipulate both components on a per-channel basis. For example, increasing the weight of the high-frequency kernel (marked in red at the bottom of Figure~\ref{fig:FADC}) leads to a stronger response for high-frequency components, which in turn increases the effective bandwidth as shown in the left of Figure~\ref{fig:adakern}, curve of $\lambda_h/\lambda_l=2$.

The FreqSelect increases the receptive field size by balancing high- and low-frequency components in the feature before feeding into dilated convolution. Since convolution tends to amplify high-frequency components~\cite{2021vision}, features after dilated convolution often exhibit a higher proportion of high-frequency components.  To capture these added high-frequency components, a small dilation rate $D$ will be favored for its large effective bandwidth, at the cost of compromised receptive field size. By suppressing the high-frequency power on the input features, our FreqSelect module is able to increase the respective field size. Specifically, As shown in Figure.\ref{fig:FADC}, FreqSelect decomposes the feature map into {\color{black} 4 frequency channels} from low to high. Then, we spatially reweights each channel with a selection map to balance frequency power, enabling FADC to effectively learn a larger receptive field.



Our experimental results in semantic segmentation show that our proposed method consistently brings improvements, thus validating the effectiveness of our approach. 
In particular, when our proposed method is applied with PIDNet, it achieves the optimal balance between inference speed and accuracy on Cityscapes, resulting in an 81.0 mIoU at 37.7 FPS. Moreover, our proposed strategy can also be integrated into deformable convolution and dilated attention, resulting in a consistent boost in performance for both segmentation and object detection tasks.
Our contributions can be summarized as follows:
\begin{itemize}
\item We conduct an in-depth exploration of dilated convolution using frequency analysis, reframing the assignment of dilation as a trade-off problem that involves balancing effective bandwidth and receptive field.
\item We introduced Frequency-Adaptive Dilated Convolution (FADC). It adopts the Adaptive Dilation Rate (AdaDR), Adaptive Kernel (AdaKern), and Frequency Selection (FreqSelect) strategies. AdaDR dynamically adjusts dilation rates in a spatially variant manner to achieve a balance between effective bandwidth and receptive field. AdaKern adaptively adjusts the kernel to fully utilize the bandwidth, and FreqSelect learns a frequency-balanced feature to encourage a large receptive field.
\item We validate our approach through comprehensive experiments in the segmentation task, consistently demonstrating its effectiveness. Furthermore, the proposed AdaKern and FreqSelect also prove to be effective when integrated with deformable convolution and dilated attention in object detection and segmentation tasks.
\end{itemize}

\section{Related work}
\noindent{\bf Content-Adaptive Networks.}
As deep learning technology advances~\cite{xiang2019global,wei2022tfpnp,fuying-2021-neurocomputing,2022Guided,fuying-2018-tcsvt,Shao_2021_ICCV,zhang2022deep}, 
the effectiveness of content-adaptive characteristics has been demonstrated by various works~\cite{2021connection, 2020dyconv, 2019pixeladaptive, 2021decoupleddyconv, 2017deformable, 2021flexconv}.
One content-adaptive strategy involves weight adjustments, which are widely employed.
Recent vision transformers~\cite{2020vit, 2021swin, 2023nat} incorporate attention mechanisms to predict input-adaptive attention values. These models have achieved significant success with large receptive, but suffer from heavy computation.

In addition to weight adjustments, ~\cite{2017deformable, 2019deformablev2, 2023internimage, 2022adcnn, 2020adaptivedilated, 2023dilated} modify the sampling grid of the convolution kernel that is closely related to our work.
Deformable convolution~\cite{2017deformable, 2019deformablev2, 2023internimage} is employed in various computer vision tasks, including object detection. It introduces $K\times K \times 2$ asymmetrical offsets for every position in the sampling grid, causing the extracted features to exhibit spatial deviations. In object detection tasks, estimated boxes are corrected through regression to mitigate these deviations.
However, in position-sensitive tasks such as semantic segmentation, where strong consistency in density and features at each location is crucial, features with spatial deviations can lead to incorrect learning.
In contrast, the proposed frequency-adaptive dilated convolution only requires one value as the dilation rate for each position. This approach necessitates fewer additional standard convolutions for computing sampling coordinates, making it lightweight. Moreover, it eliminates spatial deviations, thereby reducing the risk of erroneous learning and benefiting position-sensitive tasks.

Adaptive Dilated Convolution ~\cite{2022adcnn, 2020adaptivedilated, 2023dilated} also discards the use of globally fixed dilation. \cite{2023dilated} formulates the dilation of each point in the kernel as learned fixed weights, while \cite{2022adcnn, 2020adaptivedilated} empirically adjust the dilation rate based on the assumption that dilation values are linked to inter-layer patterns between convolution layers or the object scale.
In contrast to~\cite{2022adcnn, 2020adaptivedilated, 2023dilated}, which rely on intuitive assumptions, our proposed method is motivated by quantitative frequency analysis. Moreover, they overlook the aliasing artifacts that occur when the feature frequency exceeds the sampling rate, exposing them to a potential risk of degradation.
\vspace{+0.518mm}
\noindent{\bf Aliasing Artifacts in Neural Networks.}
The issue of aliasing artifacts in neural networks is gaining increasing attention within the computer vision community. Several studies have analyzed the aliasing artifacts resulting from insufficient sampling during downsampling in neural networks~\cite{2019makingshiftinvariant, 2020delving, 2023depthadablur, 2021wavecnet, 2021aliasingimpact}.
Others have broadened their focus to include anti-aliasing techniques in various applications, such as vision transformers~\cite{2021blending}, tiny object detection~\cite{2023antialiasingtiny}, and image generation in generative adversarial networks (GANs)~\cite{2021aliasgan}.
Regarding aliasing artifacts in dilated convolution, commonly referred to as the gridding artifact, they occur when a feature map contains higher-frequency content than the sampling rate of the dilated convolution~\cite{2017dilated}.
Previous works either empirically applied learned convolution to acquire low-pass filters for anti-aliasing~\cite{2017dilated}, employed dilated convolution with multiple dilation rates~\cite{2018understanding, 2021d3net}, or used a fully connected layer to smooth dilated convolutions~\cite{2018smoothed}. However, these methods are primarily empirically designed, involving stacking more layers, and do not explicitly handle the issue from a frequency perspective.
In contrast, our proposed method avoids gridding artifacts by dynamically adjusting the dilation rate based on local frequency. Additionally, FreqSelect contributes by suppressing high frequencies in the background or object center. This approach offers a more principled and effective solution to address aliasing artifacts.

\noindent{\bf Frequency domain learning.}
Traditional signal processing has long relied on frequency-domain analysis as a fundamental tool~\cite{1994digital, 2000digital}. Notably, these well-established methods have recently found applications in deep learning, playing pivotal roles. In this context, they are employed to examine the optimization strategies~\cite{2019fourier} and generalization capabilities~\cite{2020highfrequency} of Deep Neural Networks (DNNs).
Moreover, these frequency-domain techniques have been seamlessly integrated into DNN architectures. This integration has facilitated the learning of non-local features~\cite{2020ffc, 2021gfnet, 2021FNO, 2022AFNO, 2023adaptivefrequency} or domain-generalizable representations~\cite{2023dff}. 
Recent studies~\cite{2023spanet, 2021vision} demonstrate that capturing balanced representations of both high- and low-frequency components can enhance model performance. Therefore, 
our method provides a frequency view for dilated convolution and improves its capability to capture different frequency information.

\vspace{+0.518mm}
\section{Frequency Adaptive Dilated Convolution}
The overview of the proposed FADC is illustrated in Figure~\ref{fig:FADC}. 
In this section, we begin by introducing the AdaDR strategy, outlining how we balance bandwidth and receptive field. 
Subsequently, we delve into the details of the AdaKern and FreqSelect strategies, designed to fully leverage bandwidth and promote a large receptive field.

\subsection{Adaptive Dilation Rate}
\noindent{\bf Dilated Convolution.}
The widely-used dilated convolution can be formulated as follows:
\vspace{-2mm}
\begin{equation}
\begin{aligned}
\mathbf{Y}(p) = \sum_{i=1}^{K\times K} \mathbf{W}_i \mathbf{X}(p + \Delta p_i \times D),
\end{aligned}
\vspace{-2mm}
\end{equation}
where $\mathbf{Y}(p)$ represents the pixel value at position $p$ in the output feature map, $K$ is the kernel size, $\mathbf{W}_i$ denotes the weight parameters for the kernel, and $\mathbf{X}(p + \Delta p_i)$ represents the pixel value at the position corresponding to $p$ offset by $\Delta p_i$ in the input feature map. The variable $\Delta p_i$ represents the $i$-th location of the pre-defined grid sampling $ {(-1, -1), (-1, 0), (-1, +1), \ldots, (+1, +1)} $.
By receptive field can be enlarged by increasing the dilation rate $D$.

\vspace{+0.518mm}
\noindent{\bf Frequency analysis.}
Previous works have observed that an increased dilation leads to the degradation of frequency information capture~\cite{2017dilated, 2018understanding, 2018smoothed}. 
Specifically, increasing the dilation rate from 1 to \(D\) scales up the convolution kernel by a factor of \(D\), following the scaling property of Fourier Transforms. Consequently, the response frequency of the convolution kernel is reduced to \(\frac{1}{D}\), resulting in a shift in the frequency response from high frequency to lower frequency~\cite{1987digital, 2013digital}, as depicted in Figure~\ref{fig:intro}. 
Moreover, dilated convolution effectively operates at a sampling rate of $\frac{1}{D}$, making it unable to capture frequencies above the Nyquist frequency, \ie, half the sampling rate $\frac{1}{2D}$.

Specifically, we first transform the feature map $\mathbf{X}\in \mathbb{R}^{H\times W}$ into the frequency domain using the Discrete Fourier Transform (DFT), $\mathbf{X}_F = \mathcal{F}(\mathbf{X})$, it can be represented as
\vspace{-2mm}
\begin{equation}
\begin{aligned}
\mathbf{X}_F(u, v) = \frac{1}{HW}\sum_{h=0}^{H-1}\sum_{w=0}^{W-1}\mathbf{X}(h, w)e^{-2\pi j (uh + vw)},
\end{aligned}
\vspace{-2mm}
\end{equation}
where $\mathbf{X}_F \in \mathbb{R}^{H \times W}$ represents the output array of complex numbers from the DFT. $H$ and $W$ denote its height and width. 
$h$, $w$ indicates the coordinates of feature map $\mathbf{X}$.
The normalized frequencies in the height and width dimensions are given by $\left|u\right|$ and $\left|v\right|$.
After shifting the low frequency to the center, $u$ takes values from the set $\{-\frac{H}{2}, -\frac{H+1}{2}, \ldots, \frac{H-1}{2}\}$, and $v$ takes values from $\{-\frac{W}{2}, -\frac{W+1}{2}, \ldots, \frac{W-1}{2}\}$.
Consequently, the set of high frequencies larger than the Nyquist frequency $\mathcal{H}^{+}_D = \{(u, v) \mid \left|k \right| > \frac{{1}}{2D} \text{ or } \left|l \right| > \frac{{1}}{2D}\}$ is unable to be accurately captured, limiting its bandwidth.

\begin{figure}[tb!]
\centering
\scalebox{0.98}{
\begin{tabular}{cc}
\hspace{-2.8mm}
\includegraphics[width=0.98\linewidth]{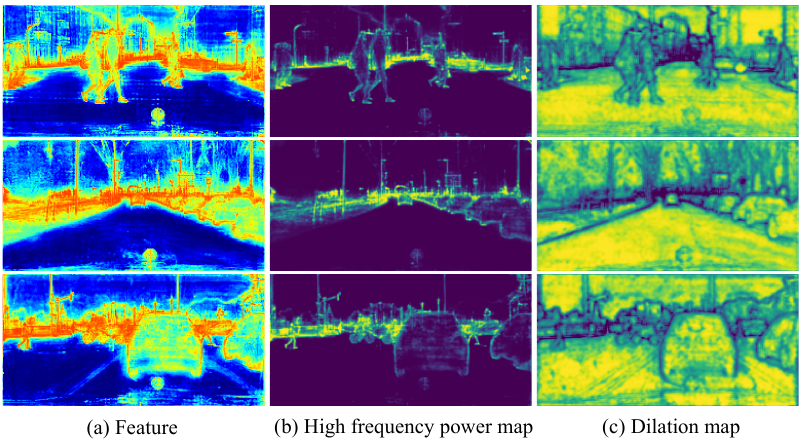}
\vspace{-5mm} 
\end{tabular}}
\caption{
Feature Visualization.
Pixels with brighter colors on the high-frequency power map indicate a higher level of high-frequency components. Brighter colors on the dilation map indicate higher dilation rates. We observe that FADC learns to assign lower dilation rates to high-frequency areas, such as the boundaries of objects, and higher dilation rates for low-frequency areas, such as the center of objects and the background.
}
\label{fig:dilation_vis}
\vspace{-5mm} 
\end{figure}

\vspace{+0.518mm}
\noindent{\bf Adaptive dilation rate.}
Building on the above analysis, the selection of the dilation rate can be viewed as a trade-off between a large receptive field and effective bandwidth. Considering that the input feature map is spatially variant, the optimal dilation for each pixel can be different. Thus, we introduce the strategy of Adaptive Dilation Rate (AdaDR) to achieve better balancing.
It assigns each pixel a different dilation rate
\vspace{-2mm}
\begin{equation}
\begin{aligned}
\mathbf{Y}(p) = \sum_{i=1}^{K\times K} \mathbf{W}_i\mathbf{X}(p + \Delta p_i \times \mathbf{\hat D}(p)).
\end{aligned}
\vspace{-2mm}
\end{equation}
$\mathbf{\hat D}(p)$ can be predicted by a convolutional layer with parameters $\theta$. 
Particularly, we incorporate a ReLU layer to ensure the non-negativity of the dilations, and we also adopt the modulation mechanism~\cite{2019deformablev2}.
It aims to maximize the receptive field and minimize the lost frequency information for each pixel. For a local feature centered at $p$ with a window size of $s$, we term it as $\mathbf{X}^{(p, s)}$. Its receptive field $\text{RF}(p)=(K - 1) \times \hat{\mathbf{D}}(p) + 1$ is positively related to $\hat{\mathbf{D}}(p)$. The frequencies in a set $\mathcal{H}^{+}_{\mathbf{\hat D(p)}}$ are unable to be captured accurately. Thus, the lost frequency information can be measured by calculating the high-frequency power $\text{HP}(p) = \sum_{\mathcal{H}^{+}_{\hat{\mathbf{D}}(p)}} |\mathbf{X}^{(p, s)}_F (u, v)|^2$. Therefore, the optimization of $\theta$ can be written as
\vspace{-2mm}
\begin{equation}
\begin{aligned}
\theta = \max_{\theta}\left(\sum \text{RF}(p) - \sum \text{HP}(p)\right).
\end{aligned}
\vspace{-2mm}
\end{equation}
However, direct optimization can be impractical due to the discrete nature of the frequency set $\mathcal{H}^{+}_{\mathbf{\hat D}(p)}$, and the fact that the calculation of $\text{HP}$ is non-differentiable. 
Consequently, we choose to optimize $\mathbf{\hat D}(p)$ directly, \ie, by increasing the dilation rate at position $p$ where the $\text{HP}(p)$ is low to encourage large receptive field and suppressing the dilation rate where $\text{HP}(p)$ is high to reduce the loss of frequency information.
To formalize this optimization, we express it as follows
\vspace{-2mm}
\begin{equation}
\begin{aligned}
\theta = \max_{\theta}\left(\sum_{p\in \text{HP}^-} \mathbf{\hat D}(p) - \sum_{p\in \text{HP}^+} \mathbf{\hat D}(p)\right).
\end{aligned}
\vspace{-2mm}
\end{equation}
Here, $\text{HP}^+$ and $\text{HP}^-$ represent pixels with the highest/lowest (\eg, 25\%) high-frequency power, \ie, the brighter/darker areas in Figure~\ref{fig:dilation_vis}{\color{red}(b)}, respectively.

\subsection{Adaptive Kernel}
AdaDR achieves a delicate equilibrium between effective bandwidth and receptive field through the individual assignment of dilation rates to each pixel, optimizing both factors collectively. The effective bandwidth, intimately connected to the convolutional kernel's weight, plays a pivotal role.
Traditional convolutional kernels learn to capture features across diverse frequency bands, which are crucial for comprehending intricate visual patterns, but they become static once trained.
To further enhance effective bandwidth, we decompose convolutional kernel parameters into low-frequency and high-frequency components before introducing dynamic weighting to adjust the frequency response.  This process only adds minor additional parameters and computational overhead.
For a static convolutional kernel, its weights $\mathbf{W}$ can be decomposed as follows
\vspace{-2mm}
\begin{equation}
\begin{aligned}
\mathbf{W} = \mathbf{\bar W} + \mathbf{\hat W}.
\end{aligned}
\vspace{-2mm}
\end{equation}
Here, $\mathbf{\bar W} = \frac{1}{K \times K}\sum_{i=1}^{K\times K}\mathbf{W}_i$ represents the kernel-wise averaged $\mathbf{W}$. It functions as a low-pass $K\times K$ mean filter, followed by a 1$\times$1 convolution with parameters defined by $\mathbf{\bar W}$.
As discussed in~\cite{2022defects}, higher mean values are more likely to attenuate the high-frequency components.
The term $\mathbf{\hat W} = \mathbf{W} - \mathbf{\bar W}$ denotes the residual part, capturing local differences and extracting the high-frequency components.
After decomposition, our AdaKern dynamical adjust both high and low frequency components  and can be formally represented as
\vspace{-2mm}
\begin{equation}
\begin{aligned}
\mathbf{W}' = \lambda_l \mathbf{\bar W} + \lambda_h \mathbf{\hat W},
\end{aligned}
\vspace{-2mm}
\end{equation}
where $\lambda_l$, $\lambda_h$ are the dynamic weights for each channel, which is predicted by a simple and lightweight global pooling + convolution layers.
Dynamically adjusting the ratio of $\frac{\lambda_l}{\lambda_h}$ based on the input context, allows the network to focus on specific frequency bands and adapt to the complexity of visual patterns in the feature.
This dynamic frequency-adaptive approach enhances the network's ability to capture both low-frequency context and high-frequency local details. This, in turn, increases the effective bandwidth, leading to improved performance in segmentation tasks that require diverse feature extraction across different frequencies.


\subsection{Frequency Selection}
As indicated by prior studies~\cite{2021vision}, conventional convolution often functions as a high-pass filter. Consequently, the resulting features tend to manifest a higher proportion of high-frequency components. This inclination leads to the adoption of smaller overall dilation rates to preserve a high effective bandwidth, unfortunately compromising the size of the receptive field. FreqSelect is devised to enhance the receptive field by balancing high- and low-frequency components in the feature representations.

Specifically, FreqSelect initially decomposes features into different frequency bands by applying distinct masks in the Fourier domain:
\vspace{-2mm}
\begin{equation}
\begin{aligned}
\mathbf{X}_{b} = \mathcal{F}^{-1}(\mathcal{M}_{b} \mathbf{X}_F),
\end{aligned}
\vspace{-2mm}
\end{equation}
where $\mathcal{F}^{-1}$ denotes the inverse Fast Fourier Transform. $\mathcal{M}_{b}$ is a binary mask designed to extract the corresponding frequency:
\vspace{-2mm}
\begin{equation}
\begin{aligned}
\color{black}
\mathcal{M}_{b}(u, v) =
\begin{cases} 
1 & \text{if } \phi_b \leq \text{max}(|u|,|v|) < \phi_{b+1} \\
0 & \text{otherwise}
\end{cases}
\end{aligned}
\vspace{-2mm}
\end{equation}
Here, $\phi_b$, $\phi_{b+1}$ are from $B+1$ predefined frequency thresholds $\{0, \phi_1, \phi_2, ..., \phi_{B-1}, \frac{1}{2}\}$. 
Subsequently, FreqSelect dynamically reweights the frequency components in different frequency bands spatially. This is formulated as:
\vspace{-2mm}
\begin{equation}
\begin{aligned}
\mathbf{\hat X}(i, j) = \sum_{b=0}^{B-1} \mathbf{A}_{b}(i, j) \mathbf{X}_{b}(i, j),
\end{aligned}
\vspace{-2mm}
\end{equation}
where $\mathbf{\hat X}(i, j)$ is the learned frequency-balanced feature after FreqSelect, and $\mathbf{A}_{b}\in \mathbb{R}^{H\times W}$ denotes the selection map for the $b$-th frequency band.
Specifically, we decompose the frequency in an octave-wise~\cite{2023spatialfrequency} manner into four frequency bands, \ie, $[0, \frac{1}{16})$, $[\frac{1}{16}, \frac{1}{8})$, $[\frac{1}{8}, \frac{1}{4})$, and $[\frac{1}{4}, \frac{1}{2}]$.

\section{Experiments}

\subsection{Experiments Settings}
\vspace{+0.518mm}
\noindent{\bf Datasets and Metrics.}
We evaluate our methods on several challenging semantic segmentation datasets, including Cityscapes~\cite{cityscapes2016} and ADE20K~\cite{ade20k}. 
We employ the mean Intersection over Union (mIoU) for semantic segmentation~\cite{fcn2015, 2023casid, 2022levelAware, chen2022consistency, chen2023semantic} and Average Precision (AP) for object detection/instance segmentation~\cite{2023lis, MaskRCNN2017, 2022hybridsupervised, 2021efficienthybrid, 2021crafting} as our evaluation metrics.

\vspace{+0.518mm}
\noindent{\bf Implement details.}
Mask2Former~\cite{2022mask2former}, PIDNet~\cite{2023pidnet}, ResNet/HorNet+{\color{black}UPerNet}, we keep the same setting with the original paper~\cite{2022mask2former, 2023pidnet, 2022hornet}.
On the COCO~\cite{lin2014microsoft} dataset, we adhere to common practices~\cite{2022hornet, 2022dilatedatt, 2023internimage} and train object detection and instance segmentation models for 12 (1$\times$ schedule) or 36 (3$\times$ schedule) epochs.
In the case of Dilated-ResNet, we substitute the dilated convolution at stage-3$\sim$4 with the proposed FADC. For PIDNet, the convolution at the bottleneck is replaced with the proposed FADC.
For ResNet, we replace the convolution at stage-2$\sim$4 with the proposed FADC.

\subsection{Main Results}
In this section, we initially assess the effectiveness of the proposed method through standard semantic segmentation benchmarks. 
Subsequently, we report results on real-time semantic segmentation. 
Finally, we seamlessly integrate the proposed method into pertinent deformable convolutions (DCNv2~\cite{2019deformablev2}) and advanced frameworks, such as DCN3-based InternImage~\cite{2023internimage}, along with incorporating dilated attention mechanisms as exemplified by DiNAT~\cite{2022dilatedatt}.

\begin{table}[t!]
\color{black}
\caption{
\small 
Results are reported on Cityscapes validation set~\cite{cityscapes2016}.
\vspace{-2.18mm}
}
\label{tab:cityscapes_deformable}
\centering
\scalebox{0.7918}{
\begin{tabular}{l|r|r|cc}
\toprule[1.28pt]
\multirow{1}{*}{Method} & \multirow{2}{*}{\#Params} & \multirow{2}{*}{\#FLOPS} &\multirow{2}{*}{mIoU} \\
\it Backbone: Dilated-ResNet-50~\cite{2017dilated} & & &\\
\midrule

PSPNet~\cite{pspnet} & 49.0M & 1427.5G & 77.8 \\
PSPNet~\cite{pspnet} + DCNv2~\cite{2019deformablev2} & +0.7M & +24.5G & 79.7 \\
\rowcolor{gray!18}
PSPNet~\cite{pspnet} + FADC (Ours) &+0.5M &+9.2G &\bf 80.4 \\
\midrule
DeepLabV3+~\cite{deeplabv3plus} & 43.6M & 1410.9G & 79.2 \\
DeepLabV3+~\cite{deeplabv3plus} + DCNv2~\cite{2019deformablev2} & +0.7M & +24.5G & 79.9 \\
\rowcolor{gray!18}
DeepLabV3+~\cite{deeplabv3plus} + FADC (Ours) &+0.5M &+9.2G&\bf 80.3 \\
\midrule
\it Backbone: Dilated-ResNet-101~\cite{2017dilated}\\
\midrule
DeepLabV3+~\cite{deeplabv3plus} + ADC~\cite{2022adcnn} & 62.8M & 2032.3G & 80.7 \\
\rowcolor{gray!18}
DeepLabV3+~\cite{deeplabv3plus} + FADC (Ours) & 63.9M & 2067.0G &\bf 81.5 \\
\midrule
\it Backbone: ResNet-50~\cite{resnet2016}\\
\midrule
Mask2Former~\cite{2022mask2former} & 44.0M & - & 79.4 \\
Mask2Former~\cite{2022mask2former} + DCNv2~\cite{2019deformablev2} & +0.9M & +7.7G & 80.4 \\
\rowcolor{gray!18}
Mask2Former~\cite{2022mask2former} + FADC (Ours) &+0.5M &+4.3G&\bf 80.6 \\
\bottomrule[1.28pt]
\end{tabular}
}
\vspace{-4.18mm}
\end{table}

\begin{table}[t!]
\color{black}
\caption{
\small 
Quantitative comparisons on semantic segmentation tasks with UPerNet~\cite{2018upernet} on the ADE20K validation set.
\vspace{-2.18mm}
}
\label{tab:ade20k}
\centering
\scalebox{0.858}{
\begin{tabular}{l|c|c|c|c}
\toprule[1.28pt]
\multirow{2}{*}{Method} & \multirow{2}{*}{\#Params} & \multirow{2}{*}{\#FLOPS} &\multicolumn{2}{c}{mIoU} \\
\cline{4-5}
& & & SS & MS \\
\midrule
ResNet-50~\cite{resnet2016} &66M &947G & 40.7 & 41.8 \\
ResNet-101~\cite{resnet2016} &85M &1029G & 42.9 & 44.0 \\
\rowcolor{gray!18}
ResNet-50-FADC (Ours)&67M &949G &\bf 44.4 & \bf 45.5 \\
\midrule
Swin-B~\cite{2021swin} & 121M & 1188G & 48.1 & 49.7 \\
NAT-B~\cite{2023nat} & 123M & 1137G & 48.5 & 49.7 \\ 
ConvNeXt-B~\cite{2022convnet} & 122M & 1170G & 49.1 & 49.9 \\
ConvNeXt-B-dcls~\cite{2023dilated} & 122M & 1170G & 49.3 & - \\
DAT-B~\cite{2022dat} & 121M & 1212G & 49.4 & 50.6 \\
DiNAT-B~\cite{2022dilatedatt} & 123M & 1137G & 49.6 & 50.4 \\
Focal-B~\cite{2022focalmodulation}& 126M & 1354G & 49.0 & 50.5 \\
InternImage-B~\cite{2023internimage} & 128M & 1185G & 50.8 & 51.3 \\ 
HorNet-B~\cite{2022hornet} & 126M & 1171G & 50.5 & 50.9 \\
\rowcolor{gray!18}
HorNet-B-FADC (Ours) & 128M & 1176G &\bf 51.1 &\bf 51.5 \\
\bottomrule[1.28pt]
\end{tabular}
}
\vspace{-5.18mm}
\end{table}

\begin{table}[t!]
\color{black}
\caption{
\small 
Comparison on Cityscapes~\cite{cityscapes2016}.
$\dagger$ indicates the models pre-trained by extra datasets.
We follow~\cite{2023pidnet} to test our method on a single RTX 3090 with a resolution of 1024$\times$2048.
\vspace{-2.18mm}
}
\label{tab:cityscapes_real_time}
\centering
\scalebox{0.768}{
\begin{tabular}{l|c|c|c|c|cc}
\toprule[1.08pt]
Model & \#Params & \#FLOPs &\#FPS & Val & Test \\
\midrule
DF2-Seg1~\cite{2019partial} & - & - & 67.2 & 75.9 & 74.8 \\
DF2-Seg2~\cite{2019partial} & - & - & 56.3 & 76.9 & 75.3 \\
\midrule
SwiftNetRN-18~\cite{2019defense} & 11.8M & 104.0G & 39.9 & 75.5 & 75.4 \\
SwiftNetRN-18 ens~\cite{2019defense}& 24.7M & 218.0G & 18.4 & - & 76.5 \\
\midrule
CABiNet~\cite{2021cabinet} & 2.64M & 12.0G & 76.5 & 76.6 & 75.9 \\
\midrule
BiSeNet(Res18)\cite{2018bisenet} & 49M & 55.3G & 65.5 & 74.8 & 74.7 \\
BiSeNetV2-L\cite{2021bisenet} & - & 118.5G & 47.3 & 75.8 & 75.3 \\
\midrule
STDC1-Seg75~\cite{2021rethinking} & - & - & 74.8 & 74.5 & 75.3 \\
STDC2-Seg75~\cite{2021rethinking} & - & - & 58.2 & 77.0 & 76.8 \\
\midrule
PP-LiteSeg-T2~\cite{2022ppliteseg} & - & - & 96.0 & 76.0 & 74.9 \\
PP-LiteSeg-B2~\cite{2022ppliteseg} & - & - & 68.2 & 78.2 & 77.5 \\
\midrule
HyperSeg-M~\cite{2021hyperseg} & 10.1M & 7.5G & 59.1 & 76.2 & 75.8 \\
HyperSeg-S~\cite{2021hyperseg} & 10.2M & 17.0G & 45.7 & 78.2 & 78.1 \\
\midrule
SFNet(DF2)\cite{2020semanticflow} & 10.53M & - & 87.6 & - & 77.8 \\
SFNet(ResNet-18)\cite{2020semanticflow} & 12.87M & 247.0G & 30.4 & - & 78.9 \\
SFNet(ResNet-18)$^\dagger$\cite{2020semanticflow} & 12.87M & 247.0G & 30.4 & - & 80.4 \\
\midrule
DDRNet-23-S\cite{2021DDRNet} & 5.7M & 36.3G & 108.1 & 77.8 & 77.4 \\
DDRNet-23~\cite{2021DDRNet} & 20.1M & 143.1G & 51.4 & 79.5 & 79.4 \\
DDRNet-39~\cite{2021DDRNet} & 32.3M & 281.2G & 30.8 & - & 80.4 \\
\midrule
PIDNet-S~\cite{2023pidnet} & 7.6M & 47.6G & 93.2 & 78.8 & 78.6 \\
PIDNet-M~\cite{2023pidnet} & 34.4M & 197.4G & 39.8 & 80.1 & 80.1 \\
PIDNet-L~\cite{2023pidnet} & 36.9M & 275.8G & 31.1 & 80.9 & 80.6 \\
\midrule
\rowcolor{gray!18}
\multirow{1}{*}{PIDNet-M-FADC (Ours)} & \multirow{1}{*}{34.6M} & \multirow{1}{*}{198.4G} & \multirow{1}{*}{37.7} & \multirow{1}{*}{\bfseries 81.0} & \multirow{1}{*}{\bfseries 80.6} \\
\bottomrule[1.08pt]
\end{tabular}
}
\vspace{-6.18mm}
\end{table}

\vspace{+0.518mm}
\noindent{\bf Standard Semantic Segmentation.}
As shown in Table~\ref{tab:cityscapes_deformable}, we compared the proposed FADC with Dilated Convolution~\cite{2017dilated}, Deformable Convolution (DCNv2)\cite{2019deformablev2}, and Adaptive Dilated Convolution (ADC)\cite{2022adcnn}.
On the widely used Cityscapes dataset~\cite{cityscapes2016}, when equipped with our FADC, the results for PSPNet, DeepLabV3+, and Mask2Former show improvements of +2.6, +1.1, and +1.2 mIoU, respectively. These enhancements outperform DCNv2 by 0.7, 0.4, and 0.2 mIoU with fewer additional computations and parameters. 
FADC also outperforms ADC, which adopts an adaptive dilation strategy, by 0.8 mIoU.
Furthermore, as demonstrated in Table~\ref{tab:ade20k} using the more challenging ADE20K dataset, FADC significantly enhances the mIoU of ResNet-50 with UPerNet by 3.7, surpassing even its heavier counterpart, ResNet-101 (44.4 \emph{vs.} 42.9).
When applied with larger HorNet-B, it leads to +0.6 gains and outperforms recent state-of-the-art methods, including Swin, ConvNeXt, RepLKNet-31L, InternImage, and DiNAT.
Notably, HorNet-B-FADC exhibits superior performance and improvement (51.1 \emph{vs.} 49.3, and +0.6 \emph{vs.} +0.2) compared to ConvNeXt-B-dcls~\cite{2023dilated}, which applies learning dilation spacing.

\vspace{+0.518mm}
\noindent{\bf Real-time Semantic Segmentation.}
Real-time semantic segmentation is crucial for applications such as autonomous vehicles~\cite{2020deepmultimoda} and robot surgery~\cite{2018automatic}. 
We further evaluate the proposed method for real-time semantic segmentation on the Cityscapes dataset~\cite{cityscapes2016} as shown in Table~\ref{tab:cityscapes_real_time}.
Equipped with FADC, our PIDNet-M achieves a mIoU of 81.0 at a frame rate of 37.7 frames per second (FPS), surpassing the performance of the heavier PIDNet-L while maintaining a faster speed (37.7 \emph{vs.} 31.1), thereby establishing a new state-of-the-art.
This demonstrates the efficiency of the proposed method.

\vspace{+0.518mm}
\noindent{\bf Integration with DCNv2, InternImage, and DiNAT.}
There exists a set of potent techniques for adjusting the sampling coordinates of convolution or attention, akin to dilated convolution. Examples include DCNv2~\cite{2019deformablev2}, InternImage~\cite{2023internimage} (a DCNv3-based model), and DiNAT~\cite{2022dilatedatt}.
DCNv2 and InternImage can be conceptualized as dynamically assigning a dilation rate to each point of the kernel. Conversely, DiNAT adjusts the sampling coordinates for calculating attention in a manner analogous to dilated convolution, thereby encountering similar challenges associated with dilation convolution.
Here, we combine the proposed AdaKern and FreqSelect with DCNv2, InternImage (DCNv3-based model), and DiNAT to assess their effectiveness. Table~\ref{tab:dcnv2_fasterrcnn} illustrates the impact of this integration. DCNv2 has previously demonstrated notable success in object detection tasks, and our proposed AdaKern and FreqSelect contribute a further enhancement of 0.9 in box AP.
Furthermore, FreqSelect enhances the performance of InternImage by 0.8 on the ADE20K dataset, and DiNAT by 0.6 in mask AP on COCO~\cite{mscoco2014}. These results serve as compelling evidence of the efficacy of our approach.

\begin{figure}[tb!]
\centering
\scalebox{0.98}{
\begin{tabular}{cc}
\includegraphics[width=0.98\linewidth]{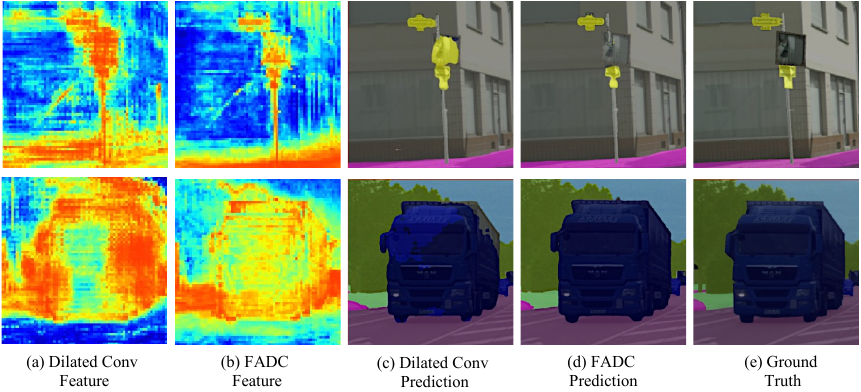}
\vspace{-2.18mm} 
\end{tabular}}
\caption{
Visualized results on Cityscape~\cite{cityscapes2016}. 
Aliasing artifacts are evident in (a), resulting in the loss of details in the representation of thin poles and truck boundaries, which leads to inferior predictions in (c). 
In contrast, our proposed FADC method in (b) demonstrates an accurate and uniform response to both thin poles and large trucks, thereby contributing to consistently accurate predictions in (d).
}
\label{fig:cityscapes}
\vspace{-2.18mm}
\end{figure}

\vspace{+0.518mm}
\noindent{\bf Visualized Results.}
We present representative visualization results in Figure~\ref{fig:cityscapes}.
The top row demonstrates that dilated convolution fails to accurately extract high-frequency information, such as the fine details of thin poles. In contrast, our proposed Frequency-Adaptive Dilated Convolution (FADC) accurately captures these details, resulting in superior predictions.
In the bottom row, it is evident that dilated convolution struggles to respond uniformly to large trucks due to an insufficient receptive field to extract local information. On the other hand, FADC uniformly responds to large trucks, leading to more consistent and accurate segmentation predictions.
These visualizations serve to illustrate the effectiveness of our proposed FADC in addressing the limitations of dilated convolution.


%

\begin{table}[tb!]
\caption{
Combining the proposed AdaKern and FreqSelect strategies with DCNv2~\cite{2019deformablev2} on the object detection task. 
All models are trained with a 1$\times$ schedule on the COCO dataset~\cite{mscoco2014}.
\vspace{-2.18mm}
}
\label{tab:dcnv2_fasterrcnn}
\centering
\scalebox{0.7528}{
\hspace{-2.18mm}
\begin{tabular}{l|c|c|ccc}
\toprule[1.28pt]
\text{Model:} Faster-RCNN~\cite{fasterRCNN2015} & \text{Param} & \text{FLOPs} &\footnotesize $\text{AP}^{box}$ &\footnotesize $\text{AP}_{50}^{box}$ &\footnotesize $\text{AP}_{75}^{box}$ \\
\midrule
ResNet-50~\cite{resnet2016}& 41.7M & 207.1G & 37.4 & 58.1 & 40.4 \\
\midrule
ResNet-50~\cite{resnet2016} + DCNv2~\cite{2019deformablev2} & +0.9M & +3.9G & 41.3 & 62.8 & 45.1  \\
\rowcolor{gray!18}
\text{+AdaKern+FreqSelect (Ours) } & +1.0M & +4.6G &\bf 42.2 &\bf 63.5 &\bf 46.2 \\
\bottomrule[1.28pt]
\end{tabular}}
\vspace{-2.18mm}
\end{table}

\begin{table}[t!]
\color{black}
\caption{
\small 
Combining the proposed FreqSelect strategies with InternImage~\cite{2023internimage} on the ADE20K validation set.
\vspace{-2.18mm}
}
\label{tab:large}
\centering
\scalebox{0.7918}{
\begin{tabularx}{0.5\textwidth}{l|c|c|c|c}
\toprule[1.28pt]
\multirow{2}{*}{Method} & \multirow{2}{*}{\#Params} & \multirow{2}{*}{\#FLOPS} &\multicolumn{2}{c}{mIoU} \\
\cline{4-5}
 & & & SS & MS \\
\midrule
Swin-T~\cite{2021swin} & 60M & 945G & 44.5 & 45.8 \\
ConvNeXt-T~\cite{2022convnet} & 60M & 939G & 46.0 & 46.7 \\
SLAK-T~\cite{2023slak} & 65M & 936G & 47.6 & - \\
\midrule
InternImage-T~\cite{2023internimage} & 59M & 944G & 47.9 & 48.1 \\
\rowcolor{gray!18}
+ FreqSelect (Ours) & 60M & 948G &\bf 48.7 &\bf 48.9 \\
\bottomrule[1.28pt]
\end{tabularx}
}
\vspace{-2.18mm}
\end{table}

\begin{table}[t!]
\centering
\caption{
Object detection and instance segmentation performance on the COCO dataset~\cite{mscoco2014} with the Mask R-CNN detector~\cite{MaskRCNN2017}.
All models are trained with a 3$\times$ schedule~\cite{2022dilatedatt, 2023internimage}.
\vspace{-2.18mm}
}
\scalebox{0.828}{
\begin{tabular}{l|c|c|c|ccc|ccc|ccc|cccccc}
\toprule[1.28pt]
Model & Params & FLOPs &\footnotesize AP$^{box}$&\footnotesize AP$^{mask}$\\
\midrule
ConvNeXt-S~\cite{2022convnet} & 348G & 70M  & 47.9  & 42.9 \\
RegionViT-B+~\cite{2021regionvit} & 307G & 93M  & 48.3 & 43.5  \\
NAT-S~\cite{2023nat} & 330G & 70M  & 48.4 & 43.2 \\
ConvNeXt-B~\cite{2022convnet} & 486G & 108M  & 48.5 & 43.5 \\
Swin-S~\cite{2021swin} & 359G & 69M  & 48.5 & 43.3 \\
Internlmage-S~\cite{2023internimage} & 340G & 69M  & 49.7 & 44.5  \\
DAT-S~\cite{2022dat} & 378G & 69M  & 49.0 & 44.0 \\
\midrule
DiNAT-S~\cite{2022dilatedatt} & 330G & 70M  & 49.3 & 43.9  \\
\rowcolor{gray!18}
+FreqSelect (Ours) & 331G & 71M  
&\bf 49.8 
&\bf 44.5 \\
\bottomrule[1.28pt]
\end{tabular}}
\label{tab:dinat}
\vspace{-2.18mm}
\end{table}

\begin{figure}[tb!]
\centering
\scalebox{0.628}{
\begin{tabular}{cc}
\hspace{-5mm}
\includegraphics[height=0.518\linewidth]{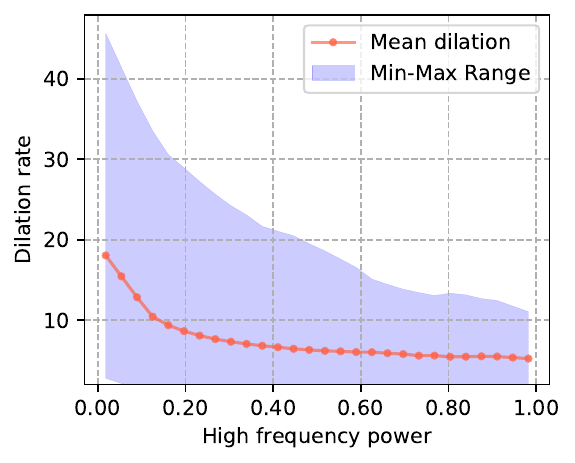} 
\hspace{-3mm}
\includegraphics[height=0.518\linewidth]{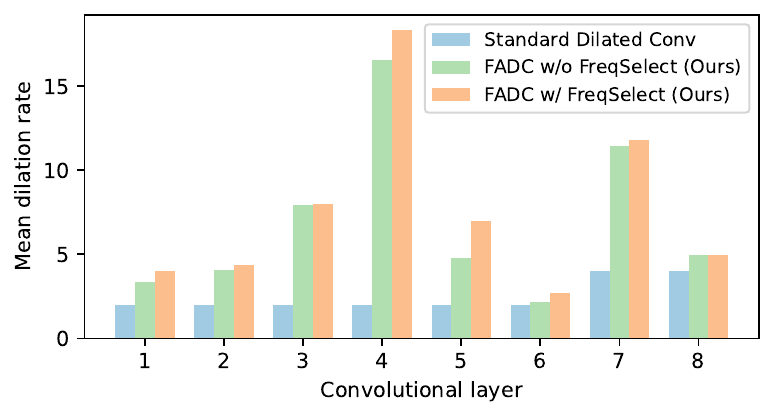} 
\vspace{-2.18mm}
\end{tabular}}
\caption{
Left: The curve illustrates the relationship between normalized high-frequency power and the predicted dilation rate. FADC, incorporating AdaDR, assigns lower dilation rates to areas with more high-frequency components, such as the boundaries of cars.
Right: Mean dilation rates at stages 3 and 4 in ResNet.
}
\label{fig:AdaDR}
\vspace{-2.18mm} 
\end{figure}

\section{Analysis and Disccusion}
We utilize dilated ResNet-50~\cite{2017dilated} as the baseline model and conduct a thorough analysis of the proposed FADC. Additional analyses are available in the supplementary material.

\begin{figure}[tb!]
\centering
\scalebox{0.998}{
\begin{tabular}{cc}
\hspace{-5mm}
\includegraphics[width=0.998\linewidth]{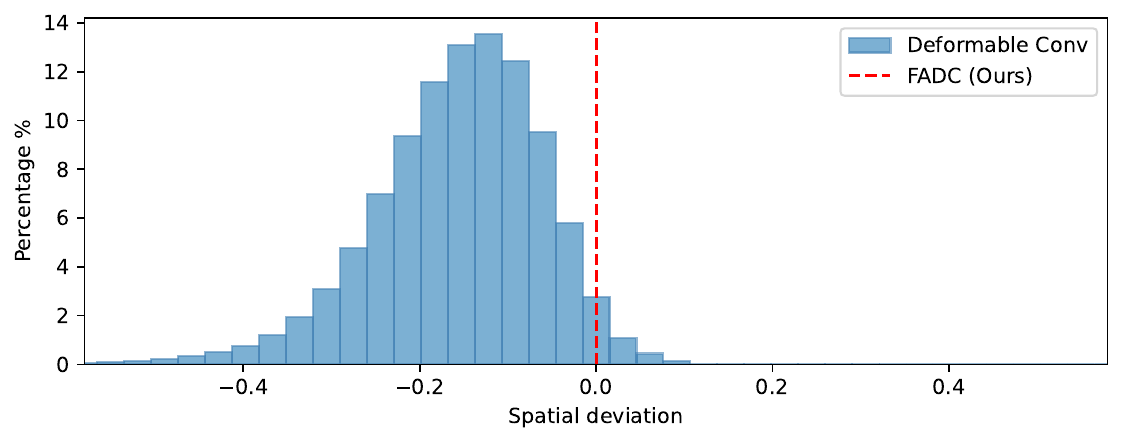} 
\vspace{-5mm}
\end{tabular}}
\caption{
Spatial deviation analysis. We illustrate a histogram of the spatial deviation between the center of the predicted sampling coordinate and corresponding pixel coordinates.
}
\label{fig:deformable}
\vspace{-2.18mm} 
\end{figure}

\vspace{+0.518mm}
\noindent{\bf Analysis of AdaDR.}
As depicted in Figure~\ref{fig:AdaDR}, AdaDR learns to predict a small dilation rate for areas with high frequencies, such as the boundaries of cars, bicycles, and persons (refer to Figure~\ref{fig:dilation_vis}{\color{red}(c)}), to maintain a high bandwidth for capturing high frequency fine details. Conversely, it assigns a larger dilation rate for smoother areas with a lower level of high frequency to expand the receptive field.
Furthermore, in comparison to deformable convolution~\cite{2017deformable, 2019deformablev2}, AdaDR avoids spatial deviation illustrated in Figure~\ref{fig:deformable}, preventing incorrect learning and benefits position-sensitive tasks.

\vspace{+0.518mm}
\noindent{\bf Analysis of AdaKern.}
Through adaptive adjustment of the ratio between high-frequency and low-frequency components in the static kernel based on input feature, AdaKern modulates the frequency response of the convolution kernel, empowering FADC to extract more high-frequency detailed information.
As depicted in the right of Figure~\ref{fig:adakern}, we perform a statistical analysis of the frequency power in the feature map. In comparison to dilated convolution, FADC extracts a greater amount of high-frequency information, crucial for capturing segmentation details, and using AdaKern further amplifies this capability.

\begin{figure}[tb!]
\centering
\scalebox{0.98}{
\begin{tabular}{cc}
\hspace{-2.18mm}
\includegraphics[width=0.98\linewidth]{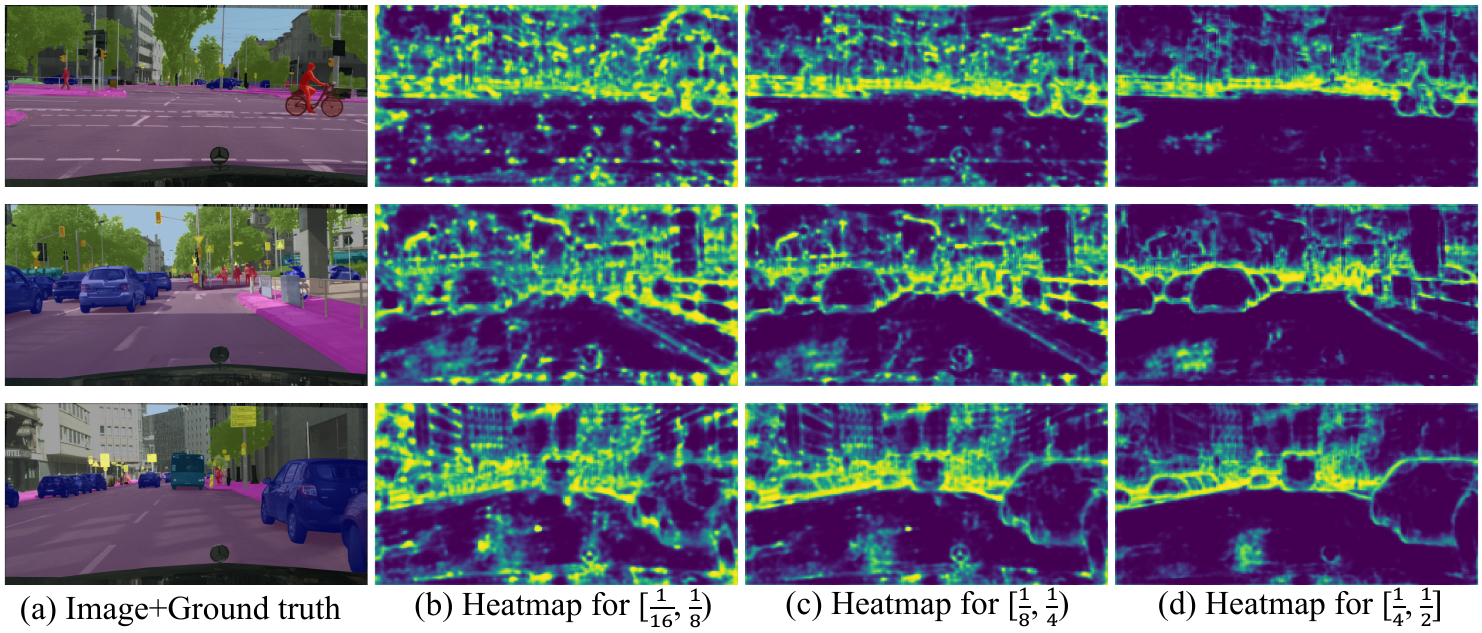}
\vspace{-3.98mm}
\end{tabular}}
\caption{
Visualization for Freqselect.
Brighter colors indicate a higher attention value.
}
\label{fig:freqselect}
\vspace{-2.18mm} 
\end{figure}

\begin{table}[tb!]
\caption{
Theoretical receptive field analysis using dilated ResNet-50~\cite{2017dilated}.
FADC considerably improves the receptive field of the entire model, and FreqSelect further enhances it.
\vspace{-2.18mm}
}
\label{tab:receptive_field}
\centering
\scalebox{0.78}{
\begin{tabular}{c|c|c|ccccccc}
\toprule[1.28pt]
\multirow{2}{*}{Conv. Type}& Standard& FADC &  FADC \\
 & Dilated Conv & (w/o FreqSelect) & (w/ FreqSelect)\\
\midrule
Receptive Field &  441 & 1007 &\bf 1100 \\
\bottomrule[1.28pt]
\end{tabular}}
\vspace{-2.18mm}
\end{table}

\begin{table}[t!]
\caption{
Average attention weights of different frequency bands.
Statistics are collected on the Cityscapes validation set~\cite{cityscapes2016}.
\vspace{-2.18mm}
}
\label{tab:freqselect}
\centering
\scalebox{0.88}{
\begin{tabular}{c|c|c|c|cccccc}
\toprule[1.28pt]
\multirow{1}{*}{Frequency band ($\times 2\pi$)} & [0, $\frac{1}{16}$) & [$\frac{1}{16}$, $\frac{1}{8}$) &  [$\frac{1}{8}$, $\frac{1}{4}$) &  [$\frac{1}{4}$, $\frac{1}{2}$] \\
\midrule
Average selection weight &  1.0 & 0.66 & 0.50 & 0.34\\
\bottomrule[1.28pt]
\end{tabular}}
\vspace{-2.18mm}
\end{table}
\vspace{+0.518mm}
\noindent{\bf Analysis of FreqSelect.}
We conduct a statistical analysis of the average weights generated by FreqSelect for different frequency bands, as presented in Table~\ref{tab:freqselect}. 
FreqSelect predicted a lower average weight for the higher frequency band, consistent with the inverse power law~\cite{2003statistics}.  
Upon visualizing the heatmap in Figure~\ref{fig:freqselect}, we noted that FreqSelect tends to assign a higher attention weight to object boundaries. This is more obvious for higher frequency bands.
It selectively suppresses high frequencies in areas that do not contribute to accurate predictions, such as the background and the center of objects. This encourages FADC to learn higher dilation rates, thereby enlarging the receptive field.

\vspace{+0.518mm}
\noindent{\bf Receptive Field.}
The importance of a large receptive field in scene understanding tasks has been emphasized~\cite{2022replknet, 2020vit}. 
Adopting the AdaDR strategy, FADC can employ a higher overall dilation rate to expand the receptive field, surpassing the widely used dilated ResNet~\cite{2017dilated} with a global fixed dilation rate, as indicated in Table~\ref{tab:receptive_field}.
Figure~\ref{fig:freqselect} visually demonstrates how FreqSelect contributes to an increased average dilation rate of FADC. By selectively weighting frequencies in the feature map, FreqSelect further encourages a higher dilation rate, ultimately resulting in an elevated receptive field, as shown in Table~\ref{tab:receptive_field}.

\vspace{+0.518mm}
\noindent{\bf Bandwidth.}
Measuring the bandwidth of a complex model is not straightforward~\cite{2021flexconv}, instead, we directly assess the frequency information in extracted features. 
In Figure~\ref{fig:adakern}, in comparison with dilated convolution, FADC increases the power in the high-frequency band of $[\frac{1}{8}, \frac{1}{4})$ and $[\frac{1}{4}, \frac{1}{2}]$.
AdaKern further enhances the power in the frequency band $[\frac{1}{4}, \frac{1}{2}]$. This indicates the extraction of more high-frequency information, demonstrating an improved bandwidth.

\vspace{+0.518mm}
\noindent{\bf Aliasing Artifacts.}
As outlined in~\cite{2017dilated, 2018understanding}, aliasing artifacts, commonly referred to as gridding artifacts, manifest when the frequency content of a feature map exceeds the sampling rate of dilated convolution, as depicted in Figure~\ref{fig:cityscapes}.
To elaborate, these artifacts occur when the frequency within the feature map surpasses the effective bandwidth of dilated convolution.
Previous studies have attempted to address this issue empirically by incorporating additional convolutional layers to learn a low-pass filter for artifact removal~\cite{2021d3net, 2018understanding} or by employing multiple dilation rates to increase the sampling rate~\cite{2021d3net, 2018understanding}. 
In contrast to these approaches, our proposed method mitigates gridding artifacts by dynamically adjusting the dilation rate based on local frequency. Furthermore, FreqSelect contributes to this by suppressing high frequencies in areas that do not contribute to accurate predictions in the background or object center.

\section{\color{black} Conclusion}
In this work, we review dilated convolution from a frequency perspective and introduce FADC to improve individual phases with three key strategies: AdaDR, AdaKern, and FreqSelect.
Diverging from the conventional approach of employing a fixed global dilation rate, AdaDR dynamically adjusts dilation rates based on local frequency components, enhancing spatial adaptability. AdaKern dynamically adjusts the ratio between low-frequency and high-frequency components in convolution weights on a per-channel basis, capturing more high-frequency information and improving overall effective bandwidth. FreqSelect balances high- and low-frequency components through spatially variant reweighting, encouraging FADC to learn larger dilations and, consequently, expanding the receptive field.
In the future, we aim to extend our quantitative frequency analysis to deformable/dilated attention. Additionally, since FADC are demonstrated to be seamlessly replace standard convolution layers in the existing architectures, we are going to design specific architecture for FADC.

\section*{Acknowledgments}
This work was supported by the National Natural Science Foundation of China (62331006, 62171038, and 62088101), the R\&D Program of Beijing Municipal Education Commission (KZ202211417048), the Fundamental Research Funds for the Central Universities, and JST Moonshot R\&D Grant Number JPMJMS2011, Japan.

{\small
\bibliographystyle{ieee_fullname}
\bibliography{./egbib.bib}
}

\end{document}